\documentclass[letterpaper]{article} 
\usepackage{aaai25}  
\usepackage{aaai25}
\usepackage{times}  
\usepackage{helvet}  
\usepackage{courier}  
\usepackage[hyphens]{url}  
\usepackage{graphicx} 
\usepackage{natbib}  
\usepackage{caption} 
\usepackage{soul}
\usepackage{graphicx}
\usepackage{booktabs}
\usepackage{float}
\usepackage{overpic}
\usepackage{subfloat}
\usepackage{subfigure}  
\usepackage[skip=0.5\baselineskip]{caption}
\usepackage{bm}
\usepackage{booktabs} 
\usepackage{amsmath}
\usepackage{amsfonts}
\usepackage{comment}
\usepackage{footmisc}
\usepackage{mathrsfs} 
\usepackage{enumitem}
\usepackage{array}
\usepackage{float}
\usepackage{subcaption}
\usepackage{subfig}
\usepackage{booktabs} 
\usepackage{comment}
\usepackage{graphicx}
\usepackage{amsmath}
\usepackage{multirow}
\usepackage{multicol}
\usepackage{threeparttable}
\usepackage{xspace}
\usepackage{color}
\usepackage{algorithmicx,algorithm}
\usepackage{algpseudocode}
\usepackage{algorithm,algpseudocode}
\usepackage{marvosym}   
\usepackage{colortbl}   
\usepackage{tcolorbox}  
\tcbuselibrary{breakable}   
\usepackage{wasysym}    
\usepackage[figuresright]{rotating} 
\usepackage{newfloat}
\usepackage{listings}
\DeclareCaptionStyle{ruled}{labelfont=normalfont,labelsep=colon,strut=off} 
\floatstyle{ruled}
\newfloat{listing}{tb}{lst}{}
\floatname{listing}{Listing}
\usepackage{xspace}

\title{SIGMA: Selective Gated Mamba for Sequential Recommendation}

\author {
    Ziwei Liu\textsuperscript{\rm 1}\equalcontrib,
    Qidong Liu\textsuperscript{\rm 1,2}\equalcontrib,
    Yejing Wang\textsuperscript{\rm 1},
    Wanyu Wang\textsuperscript{\rm 1}\thanks{Corresponding Author},
    Pengyue Jia\textsuperscript{\rm 1},
    Maolin Wang\textsuperscript{\rm 1}, \\
    Zitao Liu\textsuperscript{\rm 3},
    Yi Chang\textsuperscript{\rm 4},
    Xiangyu Zhao\textsuperscript{\rm 1}
}
\affiliations {
    \textsuperscript{\rm 1}City University of Hong Kong \\
    \textsuperscript{\rm 2}School of Auto. Science \& Engineering, MOEKLINNS Lab, Xi'an Jiaotong University \\
    \textsuperscript{\rm 3}Jinan University\\
    \textsuperscript{\rm 4}Jilin University\\
    \{ziwliu8-c, yejing.wang, wanyuwang4-c, jia.pengyue, morin.wang\}@my.cityu.edu.hk, 
    liuqidong@stu.xjtu.edu.cn, xianzhao@cityu.edu.hk, liuzitao@jnu.edu.cn, yichang@jlu.edu.cn
}

\begin{document}

\maketitle
\begin{abstract}
Sequential Recommender Systems (SRS) have emerged as a promising technique across various domains, excelling at capturing complex user preferences. Current SRS have employed transformer-based models to give the next-item prediction. However, their quadratic computational complexity often lead to notable inefficiencies, posing a significant obstacle to real-time recommendation processes. Recently, Mamba has demonstrated its exceptional effectiveness in time series prediction, delivering substantial improvements in both efficiency and effectiveness. However, directly applying Mamba to SRS poses certain challenges. Its unidirectional structure may impede the ability to capture contextual information in user-item interactions, while its instability in state estimation may hinder the ability to capture short-term patterns in interaction sequences.
To address these issues, we propose a novel framework called \textbf{\underline{S}}elect\textbf{\underline{I}}ve \textbf{\underline{G}}ated \textbf{\underline{MA}}mba for Sequential Recommendation (SIGMA). By introducing the Partially Flipped Mamba (PF-Mamba), we construct a special bi-directional structure to address the context modeling challenge. Then, to consolidate PF-Mamba's performance, we employ an input-dependent Dense Selective Gate (DS Gate) to allocate the weights of the two directions and further filter the sequential information. Moreover, for short sequence modeling, we devise a Feature Extract GRU (FE-GRU) to capture the short-term dependencies. Experimental results demonstrate that SIGMA significantly outperforms existing baselines across five real-world datasets. Our implementation code is available at \textcolor{blue}{https://github.com/Applied-Machine-Learning-Lab/SIMGA}. 
\end{abstract}
\section{Introduction} \label{sec:intro}
Over the past decade, sequential recommender systems (SRS) have demonstrated promising potential across various domains, including content streaming platforms~\cite{36,39}, e-commerce~\cite{37} and other domains~\cite{52}. To harness this potential and meet the demand for accurate next-item predictions~\cite{34,51}, an increasing number of researchers are focusing on refining existing architectures and proposing novel approaches~\cite{35,46,50}.
\begin{figure}[!t]
	\centering
	\includegraphics[width = 0.95\linewidth]{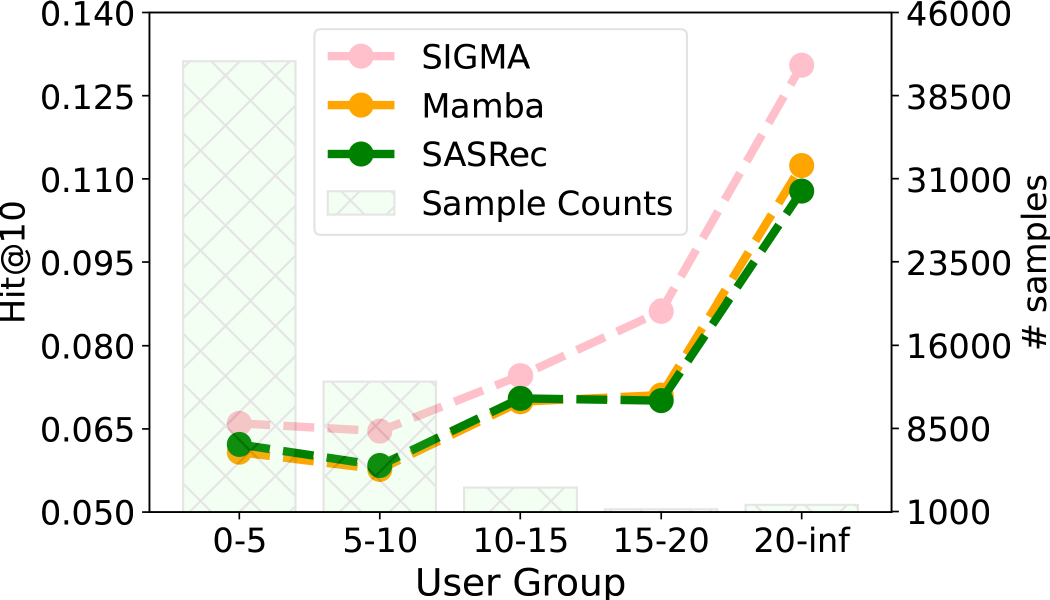}
	\caption{The illustration for long-tail user problem.}
        \label{fig:intro}
\end{figure}

Recently, Transformer-based models have emerged as the leading approaches in sequential recommendation due to their outstanding performance~\cite{9}. By leveraging the powerful self-attention mechanism~\cite{6,7}, these models have demonstrated a remarkable ability to deliver accurate predictions. However, despite their impressive performance, current transformer-based models are proven inefficient since the amount of computation grows quadratically as the length of the input sequence increases~\cite{7}. Other approaches, such as RNN-based models~\cite{11} and MLP-based models~\cite{45,26,57}, are proven to be efficient due to their linear complexity. Nevertheless, they have struggled with handling long and complex patterns~\cite{44}. All these methods above seem to have suffered from a significant trade-off between effectiveness and efficiency. 
Consequently, a specially designed State Space Model (SSM) called Mamba~\cite{1} has been proposed. By employing simple input-dependent selection on the original SSM~\cite{2,20}, it has demonstrated remarkable efficiency and effectiveness.
\begin{figure*}[!t]
	\centering
	\includegraphics[width = \linewidth]{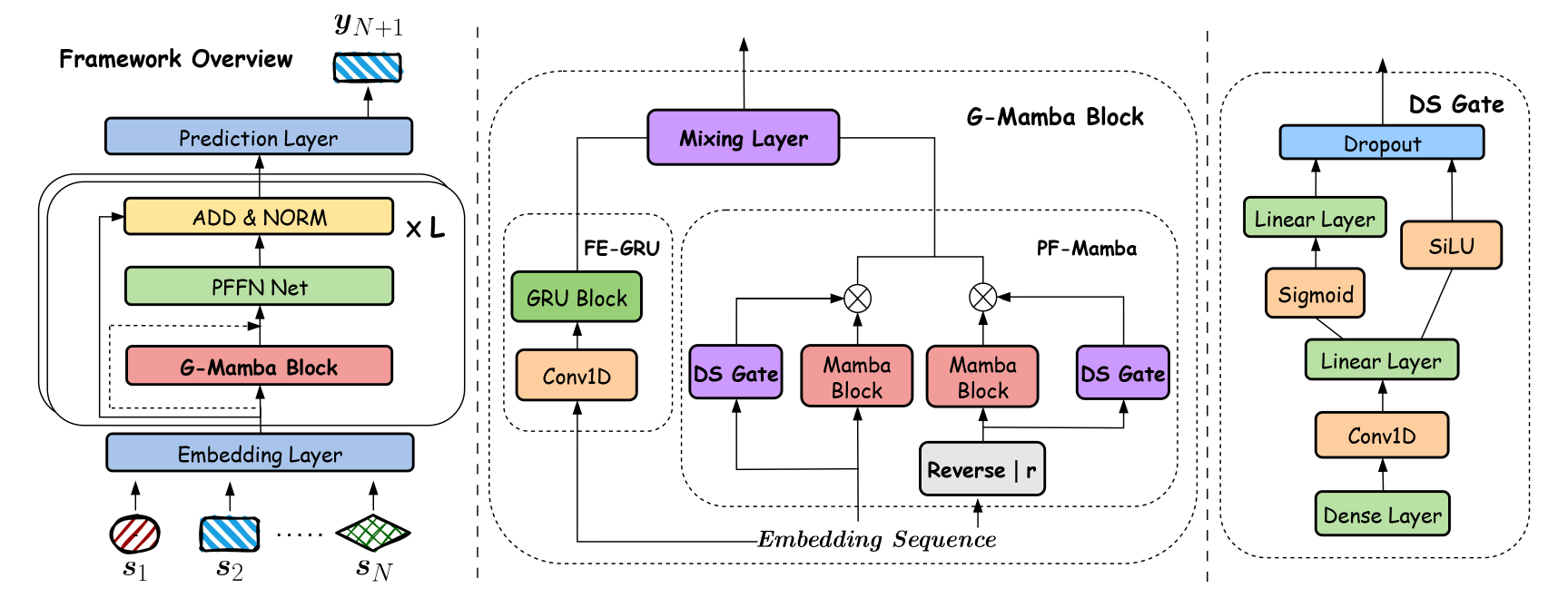}
	\caption{Framework of proposed SIGMA. The core part of this framework is the G-Mamba Block, which can directly tackle the context modeling and short sequence modeling challenges when introducing Mamba to SRS.}
        \label{fig:Overview}
\end{figure*}

However, two significant challenges hinder the direct adoption of Mamba in SRS: \begin{itemize}[leftmargin=*]
    \item \textbf{Context Modeling}: 
    While previous researches has demonstrated Mamba’s reliability in capturing sequential information~\cite{1,4}, its unidirectional architecture imposes significant limitations when applying to SRS. By only capturing users’ past behaviors, Mamba can not leverage future contextual information, potentially leading to an incomplete understanding of users' preferences ~\cite {2,15}.
    For instance, if a user consistently purchases household items but begins to show interest in sports equipment, a model that does not consider future context may struggle to recognize this shift, resulting in sub-optimal next-item predictions~\cite{16,18}.

    \item \textbf{Short Sequence Modeling}: 
    This challenge is primarily driven by the long-tail user problem, a common issue in sequential recommendation. Long-tail users refer to such users who interact with only a few items but typically receive lower-quality recommendations compared to the normal ones~\cite{10,14,60}. Furthermore, the instability in state estimation caused by limited data in short sequences~\cite{1,19,21} exacerbates this problem when Mamba is directly applied to SRS, highlighting the need for effectively modeling short sequences. For illustration, we compare two leading baselines, Mamba4Rec~\cite{2} and SASRec~\cite{8}, against our proposed framework on the Beauty dataset. As shown in Figure~\ref{fig:intro}, the histogram depicts the number of users in each group, while the line represents recommendation performance in terms of Hit@10. SASRec outperforms Mamba4Rec in the first three groups, demonstrating Mamba4Rec's exacerbation of the long-tail user problem. 

\end{itemize}
To address these challenges and better leverage Mamba's strengths, we propose an innovative framework called \textbf{\underline{S}}elect\textbf{\underline{I}}ve \textbf{\underline{G}}ated \textbf{\underline{MA}}mba for Sequential Recommendation (SIGMA). Our approach introduces the Partially Flipped Mamba (PF-Mamba), a specialized bidirectional structure that captures contextual information~\cite{2,16}. We then introduce an input-dependent Dense Selective Gate (DS Gate) to allocate the weights of the two directions and further filter the information. Additionally, we develop a Feature Extract GRU (FE-GRU) to better model short-term patterns in interaction sequences~\cite{13}, offering a possible solution to the long-tail user problem. 
Our contributions are summarized as follows: 
\begin{itemize}[leftmargin=*]
    \item We identify the limitations of Mamba when applied to SRS, attributing them to its unidirectional structure and instability in state estimation for short sequences.
    \item We introduce SIGMA, a novel framework featuring a Partially Flipped Mamba with a Dense Selective Gate and a Feature Extract GRU, which respectively address the challenges of context modeling and short sequence modeling.
    \item We validate SIGMA's performance on five public real-world datasets, demonstrating its superiority.
\end{itemize}

\section{Methodology}
In this section, we will introduce a novel framework, SIGMA, which effectively addresses the aforementioned problems by adopting PF-Mamba with a Dense Selective Gate and a Feature Extract GRU. We will first present an overview of our proposed framework; then detail the important components of our architecture; and lastly introduce how we conduct our training and inference procedures.

\subsection{Framework Overview}
In this section, we present an overview of our proposed framework in Figure~\ref{fig:Overview}. 
Firstly, we employ an embedding layer to learn the representation for input items. After getting the high-dimensional interaction representation, we propose a G-Mamba block to selectively extract the information. 
Specifically, the G-Mamba block consists of a bidirectional Mamba path and a GRU path, which respectively address challenges in context modeling and short sequence modeling. 
Then, a Position-wise Feed-Forward Network (PFFN) is adopted to improve the modeling ability of users' actions in the hidden representation. Finally, processed by the prediction layer, we can get the accurate next-item predictions.

\subsection{Embedding Layer} 
For existing SRS, It is necessary to map the sequential information in user-item interaction to a high-dimensional space~\cite{49} to effectively capture the temporal dependencies. In our framework, we choose a commonly used method for constructing the item embedding.
Here, we denote the user set as $\mathcal{U}=\left\{{u}_{1},{u}_{2},\cdots,{u}_{\left|\mathcal{U}\right|}\right\}$ and the item set as $\mathcal{V}=\left\{{v}_{1},{v}_{2},\cdots,{v}_{\left|\mathcal{V}\right|}\right\}$. So for a chronologically ordered interaction sequence, it can be expressed as $\boldsymbol{S}_{u}=\left[{s}_{1}, {s}_{2}, \cdots, {s}_{{n}_{u}}\right]$, where ${n}_{u}$ represents the length of the sequence for user $u \in \mathcal{U}$. For simplicity, we omit the mark $\left(u\right)$ in the following sections.
Regarding this interaction sequence as the input tensor, we denote $D$ as the embedding dimension and use a learnable item embedding matrix 
$\boldsymbol{E}\in \mathbb{{R}^\mathit{\left |  \mathcal{V} \right | \times D} }$ to adaptively projected ${s}_i$ into the representation $\boldsymbol{h}_i$. The whole interaction sequence can be output as: 
\begin{equation}
\label{equ:input}
\boldsymbol{H}_{0} = \left [ \boldsymbol{{h}}_{1},\boldsymbol{h}_{2},  \cdots,\boldsymbol{h}_{N}\right ] 
\end{equation}
where $N$ denotes the length of user-item interactions. 

\subsection{G-Mamba Block}
In this section, we will detail the design of our proposed G-Mamba Block. Starting with the input sequence processed by the Embedding Layer, this block introduces two paralleling paths \textit{i.e.,} PF-Mamba and FE-GRU, which respectively address the context modeling challenge and short sequence modeling challenge. Specifically, for the contextual information loss caused by the unidirectional structure of Mamba~\cite{1,18}, we introduce the Partially Flipped Mamba. It modifies the original unidirectional structure to a bi-directional one by employing a reverse block that retains the last r items while flipping the preceding items. Next, a Dense Selective Gate is proposed to properly allocate the weights of the two directions depending on the input sequence~\cite{5,25}. Additionally, for the long-tail user problem, we introduce the Feature Extract GRU to capture short-term preferences effectively~\cite{13,14}. 

\noindent \textbf{\textit{Partially Flipped Mamba.}}
This module is proposed to address the context modeling challenge by leveraging the bi-directional structure. 
Current bi-directional methods like Dual-path Mamba~\cite{16} or Vision Mamba~\cite{17} usually just flip the whole input sentence to enable the global capturing capability. Although it allows the model to have a better understanding of the context, it significantly reduces the influence of short-term patterns in interaction sequences, leading to the loss of important interest dependencies.
To address this issue, we introduce a partial flip method and integrate it with the Mamba block to construct a bi-directional structure. Followed by embedding sequence $\boldsymbol{H}_{0}$ in Equation~\eqref{equ:input}, the partially flip function adaptively reverses the first $n$ items while remaining the last $r$ items in the input tensor from $\boldsymbol{H}_{0} = \left[\boldsymbol{h}_{1}, \boldsymbol{h}_{2},\cdots,\boldsymbol{h}_{n},\boldsymbol{h}_{n+1}\cdots, \boldsymbol{h}_{N}\right]$ to $\boldsymbol{H}_{0}^{f} = \left[\boldsymbol{h}_{n},\cdots, \boldsymbol{h}_{2},\boldsymbol{h}_{1},\boldsymbol{h}_{n+1},\cdots, \boldsymbol{h}_{N}\right]$. $r$ is a pre-defined hyperparameter that equals $N-n$, which determines the range of the remaining items, \textit{i.e.,} what extent we focus on the short-term preferences. 
After processing the input sequence, we utilize two Mamba blocks to construct a bi-directional architecture and process these two sequences as follows: 
\begin{equation}
\begin{aligned}
    \boldsymbol{{M}}_{0} &= \operatorname{Mamba}\left ( \boldsymbol{H}_{0} \right ) \in \mathbb{R}^{L\times D}\\
    \boldsymbol{{M}}_{0}^{f} &= \operatorname{Mamba}\left ( \boldsymbol{H}_{0}^{f} \right ) \in \mathbb{R}^{L\times D}
\end{aligned}
\end{equation}
where ${L}$ and ${D}$ respectively represent the sequence length and hidden dimension. 
These two feature representations will then get a dot product with an input-dependent DS Gate to further learn the user preferences.
\begin{equation}
\label{equ:PFMamba}
    \hat{\boldsymbol{M}}_0 = \mathcal{G}_{1}\left ( \boldsymbol{H}_{0}\right) \cdot  \boldsymbol{M}_{0} + \mathcal{G}_{1}\left ( \boldsymbol{H}_{0}^{f}\right) \cdot \boldsymbol{M}_{0}^{f}
\end{equation}
where $\mathcal{G}_{1}$ represents the designed DS Gate and $\hat{\boldsymbol{M}}_0={\left[ \boldsymbol{m}_{1}, \boldsymbol{m}_{2}, \cdots, \boldsymbol{m}_{N}\right]}^{T}$ denotes the output from PF-Mamba. 
\noindent \textbf{\textit{Dense Selective Gate.}}
To allocate the weights of two Mamba blocks and further filter the information according to the input sequence, we design an input-dependent Dense Selective Gate. It starts with a dense layer and a Conv1d layer to extract the original sequential information from the context, which can be formalized as follows:  
\begin{equation}
    \boldsymbol{G}_{0} = \operatorname{Conv1d} \left( \boldsymbol{H}_{0} \boldsymbol{W}_{\sigma}^\mathit{(1)}+b_{\sigma}^\mathit{(1)}\right)
\end{equation}
where $\boldsymbol{H}_{0}$ is denoted as the output of embedding layer followed by Equation~\eqref{equ:input}. 
Then, we introduce a forget gate and a $\operatorname{SiLU}$ gate~\cite{5} to generate the weights from the interaction sequence:
\begin{equation}
\begin{aligned}
\boldsymbol{\delta}_{1}\left(\boldsymbol{{G}}_{0}\right) &= \boldsymbol{{G}}_{0} \boldsymbol{W}_{\delta}^{(1)}+b_{\delta}^{(1)} \\
\mathcal{G}_{0}(\boldsymbol{{G}_{0}})&=\sigma \left(\boldsymbol{\delta}_{1}\left ({\boldsymbol{G}}_{0}\right) \right)
\end{aligned}
\end{equation}
where $W_{\delta}^{(1)} \in \mathbb{R}^{D \times D}$ is the weight, $\boldsymbol{b}_{\delta}^{(1)}\in \mathbb{R}^{D}$ is bias; $\mathcal{G}_{0}$ is denoted as the symbol of forget gate; $\sigma(\cdot)$ represents the Sigmoid activation function~\cite{40}.
By employing this $\mathcal{G}_{0}$, we can control the information flow in $\boldsymbol{G}_{0}$ to selectively retain or suppress certain information~\cite{28}. 
Apart from the $\mathcal{G}_{0}$, we also employ a $\operatorname{SiLU}$ function to further improve the capability for capturing more complex patterns and features~\cite{41}. 
Therefore, We can conclude our DS Gate as follows:
\begin{equation}
\mathcal{G}_{1}(\boldsymbol{H}_{0}) = \operatorname{SiLU}\left( \boldsymbol{\delta}_{1}(\boldsymbol{G}_{0}) \right ) +  \mathcal{G}_{0}(\boldsymbol{G}_{0})   
\end{equation}
This method allows the PF-Mamba to balance two directions of the input sequence and produce a global representation. 

\noindent \textbf{\textit{Feature Extract GRU.}}
To handle Mamba's undesirable performance on short sequence modeling, we introduce one more GRU path called Feature Extract GRU in our SIGMA framework. Considering efficiency and effectiveness, we only introduce one convolution function before the GRU cell to extract and mix the features~\cite{42}.
By employing this one-dimensional convolution with a well-designed kernel size, we can aggregate and extract information from the short-term pattern of the input embedding sequence.
Then, we can extract the hidden representation by utilizing GRU's impressive capability to capture short-term dependencies. The whole processing procedure can then be formalized as follows:
\begin{equation}
\begin{aligned}
\label{equ:GRU}
\boldsymbol{C} &= \operatorname{Conv1d} \left(\boldsymbol{H}_{0} \right ) = \left[\boldsymbol{c}_{1},\boldsymbol{c}_{2},\cdots,\boldsymbol{c}_{n}\right]\\
\boldsymbol{z}_{t} & = \sigma\left(\boldsymbol{W}_{z} \cdot\left[\boldsymbol{f}_{t-1}, \boldsymbol{c}_{t}\right] + \boldsymbol{b}_{z}\right) \\
\boldsymbol{r}_{t} & = \sigma\left(\boldsymbol{W}_{r} \cdot\left[\boldsymbol{f}_{t-1}, \boldsymbol{c}_{t}\right] + \boldsymbol{b}_{r}\right) \\
\tilde{\boldsymbol{f}}_{t} & = \tanh \left(\boldsymbol{W} \cdot\left[\boldsymbol{r}_{t} \odot \boldsymbol{f}_{t-1}, \boldsymbol{c}_{t}\right] + \boldsymbol{b}\right) \\
\boldsymbol{f}_{t} & = \boldsymbol{z}_{t} \odot  \boldsymbol{f}_{t-1} + \left(1 - \boldsymbol{z}_{t}\right) \odot \tilde{\boldsymbol{f}}_{t}
\end{aligned}
\end{equation}
where $\sigma(\cdot)$ is the sigmoid activation function, $\boldsymbol{c}_{t}$ is the input of GRU module in $t^{th}$ time step, $\boldsymbol{f}_{t}$ represents the $t^{th}$ hidden states, $\boldsymbol{z}_{t}$ and $\boldsymbol{r}_{t}$ are the update gate and the reset gate, respectively. $\boldsymbol{b}_{z}$, $\boldsymbol{{b}_{r}}$, $\boldsymbol{b}$ are bias, $\boldsymbol{W}_{z}$, $\boldsymbol{W}_{r}$, $\boldsymbol{W}$ are trainable weight matrices. The final output of FE-GRU can be denoted as $\boldsymbol{F}_{0}  = \left[ \boldsymbol{f}_{1}, \boldsymbol{f}_{2}, \cdots, \boldsymbol{f}_{N}\right] \in \mathbb{R}^\mathit{ L\times D}$.

\noindent\textbf{\textit{Mixing Layer.}}
To capture user-item interactions globally and get the comprehensive hidden representation, we introduce another layer to mix the outputs of the FE-GRU and PF-Mamba for the next-item prediction. The procedure can be formalized as follows: 
\begin{equation}
\label{equ:Mix}
    \boldsymbol{Z}_{0} = {a}_{1} \boldsymbol{{M}} + {a}_{2} \boldsymbol{F}_{0} \in \mathbb{R}^\mathit{ L\times D}
\end{equation}
where ${a}_{1},{a}_{2}$ are all trainable parameters. 
Then, we employ a linear layer to capture complex relationships:
\begin{equation}
    \hat{\boldsymbol{Z}}_0 = \boldsymbol{{Z}}_{0} \boldsymbol{W}_{\delta}^{(2)}+\boldsymbol{b}_{\delta}^{(2)}
\end{equation}
where $\boldsymbol{W}_{\delta}^{(2)} \in \mathbb{R}^{D \times D}$ is the weight, $\boldsymbol{b}_{\delta}^{(2)}\in \mathbb{R}^{D}$ is bias.

\subsection{PFFN Network}
To capture the complex features, we further leverage a position-wise feed-forward network (PFFN Net)~\cite{2,8}:
\begin{equation}
\label{equ:final}
    \boldsymbol{R}_{0} = \operatorname{GELU}\left(\hat{\boldsymbol{Z}}_0 \boldsymbol{W}_{\delta}^{(3)}+b_{\delta}^{(3)} \right) \boldsymbol{W}_{\delta}^{(4)}+b_{\delta}^{(4)}
\end{equation}
where $\boldsymbol{W}_{\delta}^{(3)}\in \mathbb{R}^{D \times 4D}, \boldsymbol{W}_{\delta}^{(4)}\in \mathbb{R}^{4D \times D}, b_{\delta}^{(3)}\in \mathbb{R}^{4D}, b_{\delta}^{(4)}\in \mathbb{R}^{D}$ are parameters of two dense layers, $\boldsymbol{R}_{0}$ represents the user representation. After that, we employ a layer normalization and a residual path to stabilize the training process and ensure that the gradients flow more effectively through the network. To maintain generality, the subscript $\left(0\right)$ here only denotes that the final user representation is obtained by $1$ SIGMA layer. Actually, we can stack more such layers to better capture complex user preferences.

\subsection{Train and Inference}
In this subsection, we will present some details about the training and inference progress in our framework. As mentioned in Equation~\eqref{equ:final}, we get the mixed hidden state representation $\boldsymbol{R}_{0}$, which involves the sequential information for the first ${N}$ items. Assuming the embedding for items as $\mathbf{H}^{item}=\left[\mathbf{h}_{1}^{item},\mathbf{h}_{2}^{item},\cdots,\mathbf{h}_{K}^{item} \right] \in \mathbb{R}^{K \times D}$, where K denotes the total number of items. The details for the next-item prediction can be formalized as follows:
\begin{equation}
\begin{aligned}
    \text{logits}_{ik} = \sum_{j=1}^{d} \mathbf{R}_{ij} \cdot \mathbf{H}_{kj}^{item} \\
    {P_{ik}} = \frac{\exp(\text{logits}_{ik})}{\sum_{l=1}^{M} \exp(\text{logits}_{il})}
\end{aligned}
\end{equation}
Where $\text{logits}_{ik}$ and $\text{P}_{ik}$ respectively represent the prediction score and corresponding probability of the i-th sample for the k-th item. 
Correspondingly, we can formulate our Cross Entropy Loss (CE)~\cite{32} and minimize it as:
\begin{equation}
     \mathcal{L}_{CE} = -\frac{1}{B} \sum_{i=1}^{B} \log P_{i, y_i} 
\end{equation}
Where ${y}_{i}$ represents the actual positive sample for i-th sample and ${B}$ represents the batch size. By constantly updating the loss in each epoch, we can obtain the optimal weighting parameters and correspondingly get an accurate next-item prediction.

\begin{table}[t]
\centering
\begin{tabular}{ccccc}
\toprule[1pt]
Dataset & \# Users & \# Items & Sparsity & Avg.length \\ 
\midrule
Yelp & 82,900 & 64,210 & 99.98\% & 9.68 \\
Sports & 75,185 & 48,567 & 99.98\% & 8.07 \\
Beauty & 22,364 & 12,102 & 99.93\% & 8.88 \\
ML-1M & 6,041 & 3,417 & 95.53\% & 165.60 \\
Games & 55,145 & 17,287 & 99.94\% & 9.01\\
\bottomrule[1pt]
\end{tabular}
\caption{The statistics of datasets}
\label{tab:dataset}
\label{table:result0}
\end{table}

\begin{table*}[t]
    \centering
    \setlength{\tabcolsep}{1mm} 
    \begin{tabular*}{\textwidth}{@{\extracolsep{\fill}}c|l|cccccccc|c@{}}
        \toprule[1pt]
        \multicolumn{1}{l|}{Datasets} & Eval Metrics &  GRU4Rec &  BERT4Rec &  SASRec &  LinRec &  FEARec &  Mamba &  ECHO & \textbf{SIGMA} & Improv. \\
        \midrule
        \multirow{3}{*}{Yelp}   
        & HR@10   & 0.0441 & 0.0489 & 0.0551 & \underline{0.0579} & 0.0554 & 0.0552 & 0.0578 & \textbf{0.0629}\(^*\) & 8.82\%  \\
        & NDCG@10 & 0.0296 & 0.0317 & 0.0354 & 0.0382 & \underline{0.0391} & 0.0344 & 0.0389 & \textbf{0.0412}\(^*\) & 5.37\% \\
        & MRR@10  & 0.0218 & 0.0243 & 0.0297 & \underline{0.0322} & 0.0321 & 0.0290 & 0.0302 & \textbf{0.0346}\(^*\) & 7.45\% \\ 
        \midrule
        \multirow{3}{*}{Sports} 
        & HR@10   & 0.0523 & 0.0579 & 0.0721 & 0.0709 & \textbf{0.0746} & 0.0676 & 0.0689 & \underline{0.0735} & -1.47\% \\
        & NDCG@10 & 0.0486 & 0.0501 & 0.0546 & 0.0541 & \underline{0.0575} & 0.0563 & 0.0569 & \textbf{0.0590}\(^*\) & 2.62\%  \\
        & MRR@10  & 0.0453 & 0.0477 & 0.0513 & 0.0501 & 0.0521 & 0.0527 & \underline{0.0534} & \textbf{0.0556}\(^*\) & 4.12\%  \\ 
        \midrule
        \multirow{3}{*}{Beauty} 
        & HR@10   & 0.0612 & 0.0764 & 0.0852 & 0.0837 & \underline{0.0967} & 0.0880 & 0.0903 & \textbf{0.0986}\(^*\) & 1.96\% \\
        & NDCG@10 & 0.0334 & 0.0395 & 0.0532 & 0.0519 & 0.0530 & 0.0540 & \underline{0.0567} & \textbf{0.0604}\(^*\) & 6.53\%  \\
        & MRR@10  & 0.0242 & 0.0285 & 0.0392 & 0.0371 & 0.0410 & 0.0436 & \underline{0.0447} & \textbf{0.0488}\(^*\) & 7.83\%  \\ 
        \midrule
        \multirow{3}{*}{ML-1M}  
        & HR@10   & 0.2944 & 0.2977 & 0.2998 & 0.3102 & \underline{0.3283} & 0.3253 & 0.3239 & \textbf{0.3308}\(^*\) & 0.76\%  \\
        & NDCG@10 & 0.1652 & 0.1687 & 0.1692 & 0.1764 & 0.1843 & \underline{0.1868} & 0.1848 & \textbf{0.1906}\(^*\) & 2.03\%  \\
        & MRR@10  & 0.1252 & 0.1294 & 0.1279 & 0.1357 & \underline{0.1459} & 0.1413 & 0.1429 & \textbf{0.1479}\(^*\) & 1.37\%  \\ 
        \midrule
        \multirow{3}{*}{Games} 
        & HR@10   & 0.1484 & 0.1502 & 0.1592 & 0.1604 & \underline{0.1616} & 0.1564 & 0.1578 & \textbf{0.1627}\(^*\) & 0.68\% \\
        & NDCG@10 & 0.0964 & 0.0978 & 0.1002 & 0.1021 & 0.1032 & \underline{0.1050} & 0.1044 & \textbf{0.1088}\(^*\) & 3.62\%  \\
        & MRR@10  & 0.0735 & 0.0728 & 0.0794 & 0.0824 & 0.0843 & \underline{0.0894} & 0.0887 & \textbf{0.0924}\(^*\) & 3.36\%  \\ 
        \bottomrule[1pt]
    \end{tabular*}
    \caption{Overall performance comparison between SIGMA and other baselines. The best results are bold, and the second-best are underlined. ``*'' indicates the improvements are statistically significant (i.e., one-sided t-test with \(p<0.05\) ) over baselines.}
    \label{table:result1}
\end{table*}

\section{Experiment}
In this section, we first introduce the experiment setting. Then, we present extensive experiments to evaluate the effectiveness of SIGMA.

\subsection{Experiment Setting}
\noindent \textbf{Dataset.}
We conduct comprehensive experiments on five representative real-world datasets \textit{i.e.,} Yelp\footnote{\url{https://www.yelp.com/dataset}}, Amazon series\footnote{\url{https://cseweb.ucsd.edu/jmcauley/datasets.html\#amazon_reviews}} (Beauty, Sports and Games) and MovieLens-1M\footnote{\url{https://grouplens.org/datasets/movielens/}}.The statistics of datasets after preprocessing are shown in Table~\ref{table:result0}. For the grouped user analysis, all datasets are categorized into three subsets based on user interaction length: ``Short'' ($0-5$), ``Medium'' ($5-20$), and ``Long'' ($20+$). Additionally, we arrange user interactions sequentially by time across all datasets.

\noindent \textbf{Evaluation Metrics.}
To assess performance, we use Top-10 Hit Rate (HR@10), Top-10 Normalized Discounted Cumulative Gain (NDCG@10), and Top-10 Mean Reciprocal Rank (MRR@10) as evaluation metrics, all of which are widely used in related studies~\cite{1,16,28}. These metrics offer a comprehensive evaluation of the SRS's performance. All experimental results reported are averages from five independent runs of the framework.

\noindent \textbf{Implementation Details.}
In this section, we provide a detailed description of our framework's implementation. For GPU selection, all experiments are conducted on a single NVIDIA L4 GPU. The Adam optimizer~\cite{31} is used with a learning rate of 0.001. For a fair comparison, the embedding dimension for all tested models is set to 64. Other implementation details are the same as original papers~\cite{2,3,8}. 

\noindent \textbf{Baselines.}
To demonstrate the effectiveness and efficiency of our proposed framework, we compare SIGMA with state-of-the-art transformer-based models (\textbf{BERT4Rec}~\cite{15}, \textbf{SASRec}~\cite{8}, \textbf{LinRec}~\cite{24}, \textbf{FEARec}~\cite{23}), RNN-based models (\textbf{GRU4Rec}~\cite{11}), and SSM-based models (\textbf{Mamba4Rec}~\cite{2}, denoted as Mamba, \textbf{ECHOMamba4Rec}~\cite{3}, denoted as ECHO). %

\begin{table}[t]
    \centering
        \begin{tabular}{cccc}
            \toprule
            Dataset & Model & Inf. Time & GPU Mem. \\
            \midrule
            \multirow{6}{*}{Beauty}    
            & SASRec & 123ms & 7.58G \\
            & FEARec & 129ms & 8.11G \\
            & LinRec & \underline{72ms} & 3.08G \\
            & Mamba & \underline{72ms} & \textbf{2.58G} \\
            & ECHO & 78ms & {3.01G} \\
            & SIGMA & \textbf{68ms} & \underline{2.89G} \\ \midrule
            \multirow{6}{*}{Games}    
            & SASRec & 189ms & 7.23G \\
            & FEARec & 260ms & 7.98G \\
            & LinRec & {173ms} & 3.68G \\
            & Mamba & \underline{174ms} & {3.40G} \\
            & ECHO & 178ms & \underline{3.19G} \\
            & SIGMA & \textbf{171ms} & \textbf{3.11G} \\
            \midrule
            \multirow{6}{*}{Yelp}
            &SASRec	&443ms	&9.28G\\
	    &FEARec	&483ms	&10.01G\\
		&LinRec	&\underline{353ms}	&\underline{}{7.46G}\\
            &Mamba & 361ms & \textbf{7.32G}\\
		&ECHO	&368ms	&\underline{8.46G}\\
		&SIGMA	&\textbf{352ms}	&8.27G\\
            \bottomrule
        \end{tabular}
        \caption{Efficiency comparison of inference time per batch (ms) and GPU memory usage (GB).} 
        \label{table:result2}
\end{table}

\subsection{Overall Performance Comparison}

As shown in Table~\ref{table:result1}, we present a performance comparison on five datasets. The results show that our SIGMA framework outperforms all competing transformer-based, RNN-based, and SSM-based baselines, with significant improvements ranging from 0.76\% to 8.82\%. Such a comparison highlights the effectiveness of our unique design for combining Mamba with the sequential recommendation.

From the results, RNN-based models struggle with complex dependencies, resulting in relatively inferior performance. 
Besides, transformer-based models often show comparable performance, suggesting their powerful capacities in sequence modeling by self-attention. However, they still slightly lag behind our SIGMA because of the short sequence modeling problem they are facing and Mamba's more powerful abilities in capturing long-term dependency~\cite{4}.

In terms of the SSM-based models, we find that they also underperform our SIGMA consistently, because of the context modeling and short sequence modeling problems mentioned before. Specifically, Mamba4Rec and ECHOMamba4Rec show inferior performance in the Sports and Beauty datasets, whose average lengths are relatively shorter. Such a phenomenon emphasizes their weaknesses in long-tail users by direct adaptation of Mamba for the sequential recommendation.

\subsection{Efficiency Comparison}
In this section, we analyze the efficiency of SIGMA compared to other baselines by examining the inference time per batch and GPU memory usage during inference. The results, presented in Table~\ref{table:result2}, offer several valuable insights.
First, we can find that the Mamba-based methods, including our SIGMA, can achieve higher efficiency remarkably compared with the transformer-based methods, except for LinRec. The reason lies in the simple input-dependent selection mechanism of Mamba. 
Then, though the efficiency-specified LinRec also owns comparable efficiency, it slightly downgrades the effectiveness of the transformer. By comparison, our SIGMA can achieve a better efficiency-effectiveness trade-off.

\begin{figure}[t]
	\centering
	\includegraphics[width = \linewidth]{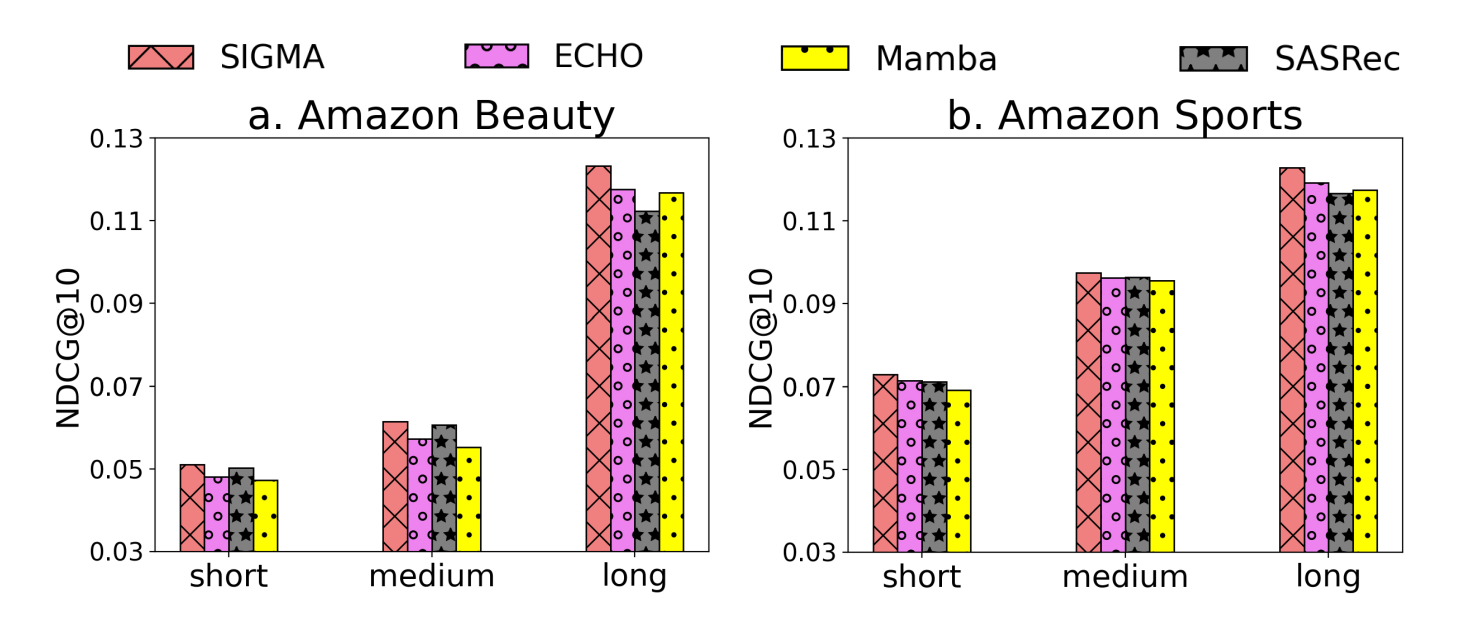}
	\caption{User group analysis on Beauty and Sports.}
        \label{fig:lt}
\end{figure}

\subsection{Grouped Users Analysis}
This section presents the recommendation quality for users with varying lengths of interaction histories, aiming to provide a deeper insight into SIGMA's effectiveness in enhancing the experience of long-tail users. We illustrate the results on Beauty and Sports in Figure~\ref{fig:lt} and find that:
\begin{itemize}[leftmargin=*]
    \item Mamba4Rec that adopts the vanilla Mamba structures for SRS presents poor performance for `short' and `medium' users. While ECHO, which designs a bi-directional modeling module for SRS, achieves slightly better results while is still worse than SASRec.
    \item Our SIGMA defeats all baselines on all groups, where FE-GRU contributes to the short-sequence modeling and PF-Mamba boosts the overall performance.
\end{itemize}

\subsection{Ablation Study}
\begin{figure}[t]
	\centering
	\includegraphics[width = \linewidth]{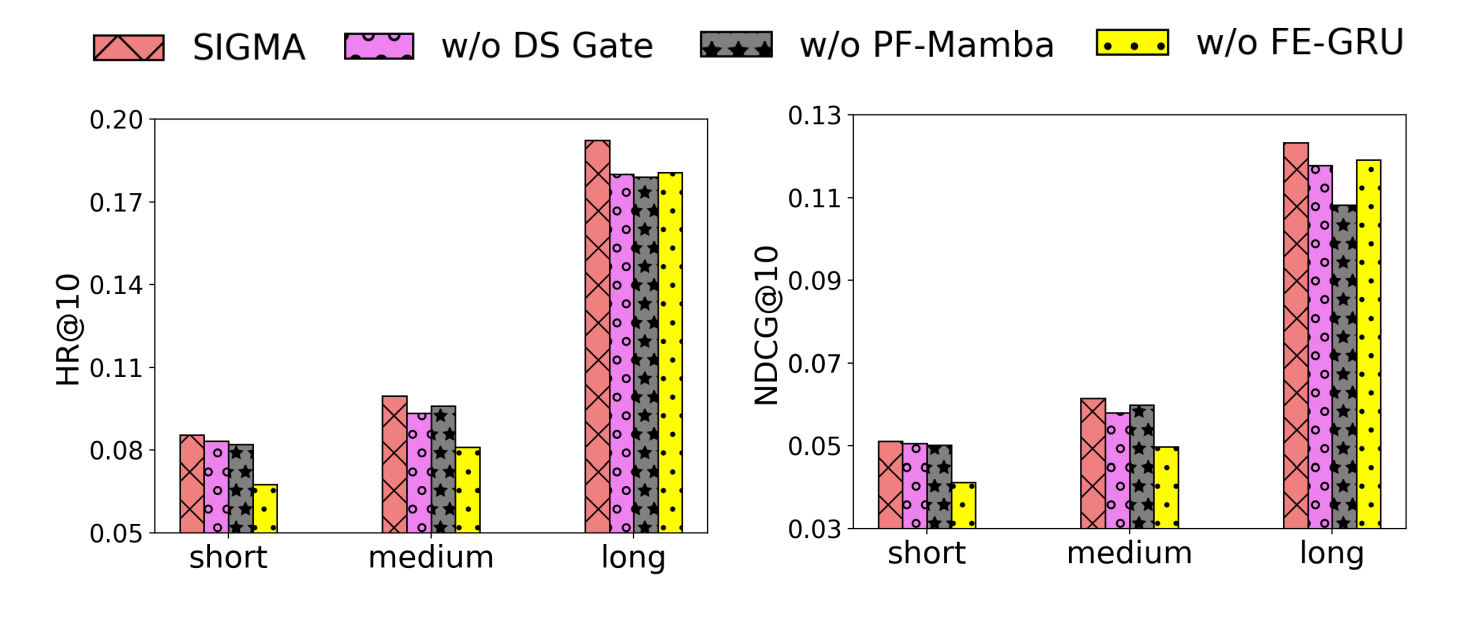}
	\caption{Ablation analysis on Beauty.}
        \label{fig:abla}
\end{figure}

In this section, we analyze the efficacy of three key components within SIGMA, including PF-Mamba (partial flipping and DS gate), and FE-GRU. We design three variants: (1) $\textit{w/o partial flipping}$: this variant uses the original interaction sequence without partial flipping; (2) $\textit{w/o DS gate}$: the second variant linearly combines the output of two Mamba blocks; (3) $\textit{w/o FE-GRU}$: this variant drops the Feature Extract GRU. We test these variants on Beauty and present results in Table~\ref{table:result3} and Figure~\ref{fig:abla}. We can conclude that: 
\begin{itemize}[leftmargin=*]
    \item With the bi-directional interaction sequences, partial flipping contributes to improving the recommendation performance for all users. 
    \item  DS gate significantly boosts the SIGMA by balancing the information from two directions. 
    \item FE-GRU is crucial for enhancing the experience of users with few interactions with strong short sequence modeling ability. And it has a huge impact on the overall performance, highlighting the importance of tackling the long-tail user problem. 
\end{itemize}

\begin{table}[t]
\centering
\setlength{\tabcolsep}{1mm} 
\begin{tabular}{cccc}
\toprule[1pt]

Model Components & HR@10 & NDCG@10 & MRR@10 \\
\midrule
Default & \textbf{0.0986} & \textbf{0.0604} & \textbf{0.0488} \\
w/o partial flipping & 0.0953 & 0.0586 & 0.0473 \\
w/o DS gate & 0.0954 & 0.0571 & 0.0455 \\
w/o FE-GRU & 0.0750 & 0.0470 & 0.0382 \\
\bottomrule[1pt]
\end{tabular}
\caption{Ablation study on Beauty.}
\label{table:result3}
\end{table}

\subsection{Hyperparameter Analysis}
\begin{figure}[t]
	\centering
	\includegraphics[width = \linewidth]{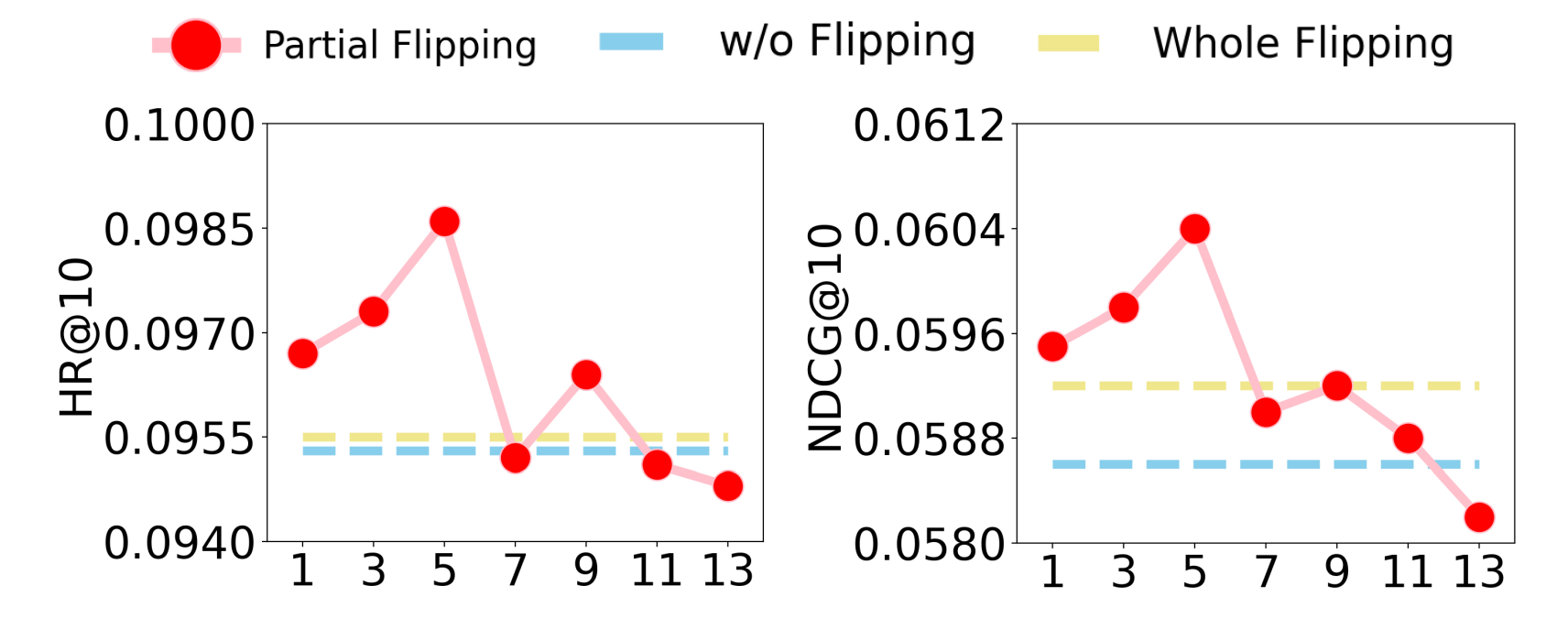}
	\caption{Parameter study for $r$ on Beauty.}
        \label{fig:hyper}
\end{figure}

In this section, we conduct experiments on Beauty to analyze the influence of two significant hyperparameters: (i) $r$, the remaining range in the partial flipping method; (ii) $L$, the number of stacked SIGMA layers. The results are respectively visualized in Figure~\ref{fig:hyper} and Table~\ref{table:result4}.

From Figure~\ref{fig:hyper}, we can find that our proposed SIGMA framework achieves the best results when $r=5$, offering two valuable insights as follows: (i) when $r$ is relatively large ($r=N$ represents ``w/o flipping" ), it is challenging for SIGMA to leverage the limited bi-directional information ($N-r$ items are flipped); (ii) when $r$ is relatively small ($r=0$ represents ``whole flipping" ), users may lose the short-term preference due to the exceeding flipping range, which is reflected as a varied Hit@10 and NDCG@10 performance in Figure~\ref{fig:hyper}. These phenomenons justify the significance of partial flipping with a proper $r$, defending the effectiveness of SIGMA.

\begin{table}[t]
	\centering
	\begin{tabular}{ccccc}
		\toprule[1pt]
		\#layers	&	HR@10 &	NDCG@10	 &Inf~Time & GPU~Mem\\
		\midrule
		$1$&		0.0986&	0.0604&	 66ms& 3.03G\\
            $2$&		0.0994&	0.0611&	 122ms& 4.30G\\
		$4$&		0.0963&	0.0589& 227ms& 6.83G\\
		\bottomrule[1pt]
	\end{tabular}
    \caption{Parameter study for $\mathit{L}$ on Beauty.}
        \label{table:result4}
\end{table} 

From Table~\ref{table:result4}, we observe that increasing the number of SIGMA layers does not guarantee the improvement of recommendation performance, but significantly impairs the inference efficiency, which can be attributed to the overfitting problem of multiple SIGMA layers. In addition, it is noteworthy that the performance of a single SIGMA layer is very close to the optimal one, indicating the strong modeling ability and superior efficiency of SIGMA.

\subsection{Case Study}
\begin{figure}[t]
	\centering
	\includegraphics[width = \linewidth]{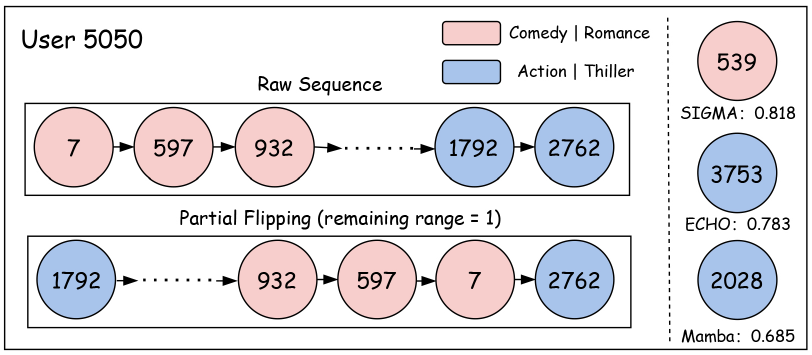}
	\caption{Case study for User-$5050$ in ML-1M.}
        \label{fig:case}
\end{figure}

In this section, we leverage a specific example in ML-1M to illustrate the effectiveness of partial flipping in SIGMA. Specifically, we choose a user (ID: 5050) and present the interaction sequence before and after the partial flipping in the left part of Figure~\ref{fig:case}. With $r=1$, only the last item $2762$ remained at the original position, and other items are flipped. From this example, we can find that this user prefers comedy and romance movies (pink balls), as well as action and thriller movies (blue balls). Without the flipping, baselines focus on the most recent interactions on action and thriller movies and provide incorrect recommendations of the same genres (movie 3753 and 2028). While our SIGMA, with PF-Mamba, notices the previous preference for comedy and romance movies, makes the accurate recommendation of movie 539. Furthermore, we also present the overall performance for user-5050 in Table~\ref{table:result5}, where SIGMA significantly defeats baselines.

\begin{table}[t]
	\centering
	\begin{tabular}{cccc}
		\toprule[1pt]
		Methods	&	HR@10 &	NDCG@10	&MRR@10 \\
		\midrule
		Mamba&		0.2998&	0.1776&	0.1391\\
            ECHO&		0.3041&	0.1799&	0.1403\\
		SIGMA&	\textbf{0.3139}& \textbf{0.1825}& \textbf{0.1423}\\
            \midrule
            Improv.& 3.22\% & 1.44\% & 1.42\%\\
		\bottomrule[1pt]
	\end{tabular}
    \caption{Performance comparison on User-$5050$.}
        \label{table:result5}
\end{table}
\section{Related Work}
\subsection{Sequential Recommendation} 
Advancements in deep learning have transformed recommendation systems, making them more personalized and accurate in next-item prediction~\cite{58,59,61}. Early sequential recommendation frameworks have adopted CNNs and RNNs to capture users' preferences but faced issues like catastrophic forgetting when dealing with long-term dependencies~\cite{9,10}. Then, the transformer-based models have emerged as powerful methods with their self-attention mechanism, significantly improving performance by selectively capturing the complex user-item interactions~\cite{47}. However, they have suffered from inefficiency due to the quadratic computational complexity~\cite{7}. Therefore, to address the trade-off between effectiveness and efficiency, we propose SIGMA, a novel framework that achieves remarkable performance.

\subsection{Selective State Space Model}
Currently, SSM-based models have been proven effective in time-series prediction due to their ability to capture the hidden dynamics~\cite{19,20}. To further address the issues of catastrophic forgetting and long-term dependency in sequential processing, a special SSM called Mamba was introduced. Attributing to its unique selectivity~\cite{1}, Mamba shows remarkable performance without leveraging any sequence denoising methods~\cite{53,54,55} or feature selecting methods~\cite{56}, even when addressing long sequences~\cite{4}. However, it still suffers from some challenges when adopted in the realm of recommendation \textit{i.e.,} context modeling and short sequence modeling, which are mainly caused by Mamba's original structure and the inflexibility in hidden state transferring. Correspondingly, we introduce a special bi-directional module called Partially Flipped Mamba and a Feature Extract GRU in our SIGMA framework, which somewhat address these problems and explores a novel way to leverage Mamba in SRS.
\section{Conclusion}
In this paper, we analyze the challenges of applying Mamba to SRS and propose a novel framework, SIGMA, to address these challenges. We introduce a bidirectional PF-Mamba, featuring a well-designed DS gate, to allocate the weights of each direction and address the context modeling challenge, enabling our framework to leverage information from both past and future user-item interactions. Furthermore, to address the challenge of short sequence modeling, we propose FE-GRU to enhance the hidden representations for interaction sequences, mitigating the impact of long-tail users to some extent. Finally, we conduct extensive experiments on five real-world datasets, verifying SIGMA's superiority and validating the effectiveness of each module.

\section*{Acknowledgements}
This research was partially supported by Research Impact Fund (No.R1015-23), APRC - CityU New Research Initiatives (No.9610565, Start-up Grant for New Faculty of CityU), CityU - HKIDS Early Career Research Grant (No.9360163), Hong Kong ITC Innovation and Technology Fund Midstream Research Programme for Universities Project (No.ITS/034/22MS), Hong Kong Environmental and Conservation Fund (No. 88/2022), and SIRG - CityU Strategic Interdisciplinary Research Grant (No.7020046), Huawei (Huawei Innovation Research Program), Tencent (CCF-Tencent Open Fund, Tencent Rhino-Bird Focused Research Program), Ant Group (CCF-Ant Research Fund, Ant Group Research Fund), Alibaba (CCF-Alimama Tech Kangaroo Fund No. 2024002), CCF-BaiChuan-Ebtech Foundation Model Fund, and Kuaishou.

\bibliography{8Reference}

\begin{thebibliography}{54}
\providecommand{\natexlab}[1]{#1}

\bibitem[{De et~al.(2024)De, Smith, Fernando, Botev, Cristian-Muraru, Gu, Haroun, Berrada, Chen, Srinivasan et~al.}]{28}
De, S.; Smith, S.~L.; Fernando, A.; Botev, A.; Cristian-Muraru, G.; Gu, A.; Haroun, R.; Berrada, L.; Chen, Y.; Srinivasan, S.; et~al. 2024.
\newblock Griffin: Mixing gated linear recurrences with local attention for efficient language models.
\newblock \emph{arXiv preprint arXiv:2402.19427}.

\bibitem[{de~Souza Pereira~Moreira et~al.(2021)de~Souza Pereira~Moreira, Rabhi, Lee, Ak, and Oldridge}]{9}
de~Souza Pereira~Moreira, G.; Rabhi, S.; Lee, J.~M.; Ak, R.; and Oldridge, E. 2021.
\newblock Transformers4rec: Bridging the gap between nlp and sequential/session-based recommendation.
\newblock In \emph{Proceedings of the 15th ACM conference on recommender systems}, 143--153.

\bibitem[{Du et~al.(2023)Du, Yuan, Zhao, Qu, Zhuang, Liu, Liu, and Sheng}]{23}
Du, X.; Yuan, H.; Zhao, P.; Qu, J.; Zhuang, F.; Liu, G.; Liu, Y.; and Sheng, V.~S. 2023.
\newblock Frequency enhanced hybrid attention network for sequential recommendation.
\newblock In \emph{Proceedings of the 46th International ACM SIGIR Conference on Research and Development in Information Retrieval}, 78--88.

\bibitem[{Fang et~al.(2020)Fang, Zhang, Shu, and Guo}]{34}
Fang, H.; Zhang, D.; Shu, Y.; and Guo, G. 2020.
\newblock Deep learning for sequential recommendation: Algorithms, influential factors, and evaluations.
\newblock \emph{ACM Transactions on Information Systems (TOIS)}, 39(1): 1--42.

\bibitem[{Gao et~al.(2024)Gao, Zhao, Li, Zhao, Wu, Guo, Liu, and Yin}]{26}
Gao, J.; Zhao, X.; Li, M.; Zhao, M.; Wu, R.; Guo, R.; Liu, Y.; and Yin, D. 2024.
\newblock SMLP4Rec: An Efficient all-MLP Architecture for Sequential Recommendations.
\newblock \emph{ACM Transactions on Information Systems}, 42(3): 1--23.

\bibitem[{Gu and Dao(2023)}]{1}
Gu, A.; and Dao, T. 2023.
\newblock Mamba: Linear-time sequence modeling with selective state spaces.
\newblock \emph{arXiv preprint arXiv:2312.00752}.

\bibitem[{Hamilton(1994)}]{20}
Hamilton, J.~D. 1994.
\newblock State-space models.
\newblock \emph{Handbook of econometrics}, 4: 3039--3080.

\bibitem[{He et~al.(2018)He, Li, Xu, and Zheng}]{40}
He, J.; Li, L.; Xu, J.; and Zheng, C. 2018.
\newblock ReLU deep neural networks and linear finite elements.
\newblock \emph{arXiv preprint arXiv:1807.03973}.

\bibitem[{Hidasi et~al.(2015)Hidasi, Karatzoglou, Baltrunas, and Tikk}]{13}
Hidasi, B.; Karatzoglou, A.; Baltrunas, L.; and Tikk, D. 2015.
\newblock Session-based recommendations with recurrent neural networks.
\newblock \emph{arXiv preprint arXiv:1511.06939}.

\bibitem[{Jannach and Ludewig(2017)}]{11}
Jannach, D.; and Ludewig, M. 2017.
\newblock When recurrent neural networks meet the neighborhood for session-based recommendation.
\newblock In \emph{Proceedings of the eleventh ACM conference on recommender systems}, 306--310.

\bibitem[{Jiang, Han, and Mesgarani(2024)}]{16}
Jiang, X.; Han, C.; and Mesgarani, N. 2024.
\newblock Dual-path mamba: Short and long-term bidirectional selective structured state space models for speech separation.
\newblock \emph{arXiv preprint arXiv:2403.18257}.

\bibitem[{Kang and McAuley(2018)}]{8}
Kang, W.-C.; and McAuley, J. 2018.
\newblock Self-attentive sequential recommendation.
\newblock In \emph{2018 IEEE international conference on data mining (ICDM)}, 197--206. IEEE.

\bibitem[{Keles, Wijewardena, and Hegde(2023)}]{7}
Keles, F.~D.; Wijewardena, P.~M.; and Hegde, C. 2023.
\newblock On the computational complexity of self-attention.
\newblock In \emph{International Conference on Algorithmic Learning Theory}, 597--619. PMLR.

\bibitem[{Kim et~al.(2019{\natexlab{a}})Kim, Kim, Park, and Yu}]{10}
Kim, Y.; Kim, K.; Park, C.; and Yu, H. 2019{\natexlab{a}}.
\newblock Sequential and Diverse Recommendation with Long Tail.
\newblock In \emph{IJCAI}, volume~19, 2740--2746.

\bibitem[{Kim et~al.(2019{\natexlab{b}})Kim, Kim, Park, and Yu}]{14}
Kim, Y.; Kim, K.; Park, C.; and Yu, H. 2019{\natexlab{b}}.
\newblock Sequential and Diverse Recommendation with Long Tail.
\newblock In \emph{IJCAI}, volume~19, 2740--2746.

\bibitem[{Kingma and Ba(2014)}]{31}
Kingma, D.~P.; and Ba, J. 2014.
\newblock Adam: A method for stochastic optimization.
\newblock \emph{arXiv preprint arXiv:1412.6980}.

\bibitem[{Kweon, Kang, and Yu(2021)}]{18}
Kweon, W.; Kang, S.; and Yu, H. 2021.
\newblock Bidirectional distillation for top-K recommender system.
\newblock In \emph{Proceedings of the Web Conference 2021}, 3861--3871.

\bibitem[{Li et~al.(2023{\natexlab{a}})Li, Wang, Liu, Zhao, Wang, Wang, Zou, Fan, and Li}]{47}
Li, C.; Wang, Y.; Liu, Q.; Zhao, X.; Wang, W.; Wang, Y.; Zou, L.; Fan, W.; and Li, Q. 2023{\natexlab{a}}.
\newblock STRec: Sparse transformer for sequential recommendations.
\newblock In \emph{Proceedings of the 17th ACM Conference on Recommender Systems}, 101--111.

\bibitem[{Li et~al.(2017)Li, Ren, Chen, Ren, Lian, and Ma}]{12}
Li, J.; Ren, P.; Chen, Z.; Ren, Z.; Lian, T.; and Ma, J. 2017.
\newblock Neural attentive session-based recommendation.
\newblock In \emph{Proceedings of the 2017 ACM on Conference on Information and Knowledge Management}, 1419--1428.

\bibitem[{Li et~al.(2023{\natexlab{b}})Li, Zhang, Zhao, Wang, Zhao, Wu, and Guo}]{45}
Li, M.; Zhang, Z.; Zhao, X.; Wang, W.; Zhao, M.; Wu, R.; and Guo, R. 2023{\natexlab{b}}.
\newblock Automlp: Automated mlp for sequential recommendations.
\newblock In \emph{Proceedings of the ACM Web Conference 2023}, 1190--1198.

\bibitem[{Li et~al.(2022)Li, Qiu, Zhao, Wang, Zhang, Xing, and Wu}]{52}
Li, X.; Qiu, Z.; Zhao, X.; Wang, Z.; Zhang, Y.; Xing, C.; and Wu, X. 2022.
\newblock Gromov-wasserstein guided representation learning for cross-domain recommendation.
\newblock In \emph{Proceedings of the 31st ACM International Conference on Information \& Knowledge Management}, 1199--1208.

\bibitem[{Liang et~al.(2023)Liang, Zhao, Li, Zhang, Wang, Liu, and Liu}]{57}
Liang, J.; Zhao, X.; Li, M.; Zhang, Z.; Wang, W.; Liu, H.; and Liu, Z. 2023.
\newblock Mmmlp: Multi-modal multilayer perceptron for sequential recommendations.
\newblock In \emph{Proceedings of the ACM Web Conference 2023}, 1109--1117.

\bibitem[{Lin et~al.(2022)Lin, Zhao, Wang, Xu, and Wu}]{56}
Lin, W.; Zhao, X.; Wang, Y.; Xu, T.; and Wu, X. 2022.
\newblock AdaFS: Adaptive feature selection in deep recommender system.
\newblock In \emph{Proceedings of the 28th ACM SIGKDD Conference on Knowledge Discovery and Data Mining}, 3309--3317.

\bibitem[{Lin et~al.(2023)Lin, Zhao, Wang, Zhu, and Wang}]{55}
Lin, W.; Zhao, X.; Wang, Y.; Zhu, Y.; and Wang, W. 2023.
\newblock Autodenoise: Automatic data instance denoising for recommendations.
\newblock In \emph{Proceedings of the ACM Web Conference 2023}, 1003--1011.

\bibitem[{Liu et~al.(2024{\natexlab{a}})Liu, Lin, Wang, Liu, and Caverlee}]{2}
Liu, C.; Lin, J.; Wang, J.; Liu, H.; and Caverlee, J. 2024{\natexlab{a}}.
\newblock Mamba4rec: Towards efficient sequential recommendation with selective state space models.
\newblock \emph{arXiv preprint arXiv:2403.03900}.

\bibitem[{Liu et~al.(2023{\natexlab{a}})Liu, Cai, Zhang, Zhao, Gao, Wang, Lv, Fan, Wang, He et~al.}]{24}
Liu, L.; Cai, L.; Zhang, C.; Zhao, X.; Gao, J.; Wang, W.; Lv, Y.; Fan, W.; Wang, Y.; He, M.; et~al. 2023{\natexlab{a}}.
\newblock Linrec: Linear attention mechanism for long-term sequential recommender systems.
\newblock In \emph{Proceedings of the 46th International ACM SIGIR Conference on Research and Development in Information Retrieval}, 289--299.

\bibitem[{Liu et~al.(2024{\natexlab{b}})Liu, Hu, Xiao, Zhao, Gao, Wang, Li, and Tang}]{46}
Liu, Q.; Hu, J.; Xiao, Y.; Zhao, X.; Gao, J.; Wang, W.; Li, Q.; and Tang, J. 2024{\natexlab{b}}.
\newblock Multimodal recommender systems: A survey.
\newblock \emph{ACM Computing Surveys}, 57(2): 1--17.

\bibitem[{Liu et~al.(2024{\natexlab{c}})Liu, Wu, Wang, Wang, Zhu, Zhao, Tian, and Zheng}]{61}
Liu, Q.; Wu, X.; Wang, W.; Wang, Y.; Zhu, Y.; Zhao, X.; Tian, F.; and Zheng, Y. 2024{\natexlab{c}}.
\newblock Large language model empowered embedding generator for sequential recommendation.
\newblock \emph{arXiv preprint arXiv:2409.19925}.

\bibitem[{Liu et~al.(2024{\natexlab{d}})Liu, Wu, Zhao, Wang, Zhang, Tian, and Zheng}]{60}
Liu, Q.; Wu, X.; Zhao, X.; Wang, Y.; Zhang, Z.; Tian, F.; and Zheng, Y. 2024{\natexlab{d}}.
\newblock Large Language Models Enhanced Sequential Recommendation for Long-tail User and Item.
\newblock \emph{arXiv preprint arXiv:2405.20646}.

\bibitem[{Liu et~al.(2023{\natexlab{b}})Liu, Cai, Sun, Wang, Jiang, Zheng, Jiang, Gai, Zhao, and Zhang}]{58}
Liu, S.; Cai, Q.; Sun, B.; Wang, Y.; Jiang, J.; Zheng, D.; Jiang, P.; Gai, K.; Zhao, X.; and Zhang, Y. 2023{\natexlab{b}}.
\newblock Exploration and regularization of the latent action space in recommendation.
\newblock In \emph{Proceedings of the ACM Web Conference 2023}, 833--844.

\bibitem[{Liu et~al.(2023{\natexlab{c}})Liu, Tian, Cai, Zhao, Gao, Liu, Chen, He, Zheng, Jiang et~al.}]{51}
Liu, Z.; Tian, J.; Cai, Q.; Zhao, X.; Gao, J.; Liu, S.; Chen, D.; He, T.; Zheng, D.; Jiang, P.; et~al. 2023{\natexlab{c}}.
\newblock Multi-task recommendations with reinforcement learning.
\newblock In \emph{Proceedings of the ACM Web Conference 2023}, 1273--1282.

\bibitem[{Nwankpa et~al.(2018)Nwankpa, Ijomah, Gachagan, and Marshall}]{41}
Nwankpa, C.; Ijomah, W.; Gachagan, A.; and Marshall, S. 2018.
\newblock Activation functions: Comparison of trends in practice and research for deep learning.
\newblock \emph{arXiv preprint arXiv:1811.03378}.

\bibitem[{Qin, Yang, and Zhong(2024)}]{5}
Qin, Z.; Yang, S.; and Zhong, Y. 2024.
\newblock Hierarchically gated recurrent neural network for sequence modeling.
\newblock \emph{Advances in Neural Information Processing Systems}, 36.

\bibitem[{Smith, Warrington, and Linderman(2022)}]{19}
Smith, J.~T.; Warrington, A.; and Linderman, S.~W. 2022.
\newblock Simplified state space layers for sequence modeling.
\newblock \emph{arXiv preprint arXiv:2208.04933}.

\bibitem[{Song et~al.(2022)Song, Chen, Zhao, Guo, and Tang}]{36}
Song, F.; Chen, B.; Zhao, X.; Guo, H.; and Tang, R. 2022.
\newblock Autoassign: Automatic shared embedding assignment in streaming recommendation.
\newblock In \emph{2022 IEEE International Conference on Data Mining (ICDM)}, 458--467. IEEE.

\bibitem[{Sun et~al.(2019)Sun, Liu, Wu, Pei, Lin, Ou, and Jiang}]{15}
Sun, F.; Liu, J.; Wu, J.; Pei, C.; Lin, X.; Ou, W.; and Jiang, P. 2019.
\newblock BERT4Rec: Sequential recommendation with bidirectional encoder representations from transformer.
\newblock In \emph{Proceedings of the 28th ACM international conference on information and knowledge management}, 1441--1450.

\bibitem[{Vaswani et~al.(2017)Vaswani, Shazeer, Parmar, Uszkoreit, Jones, Gomez, Kaiser, and Polosukhin}]{6}
Vaswani, A.; Shazeer, N.; Parmar, N.; Uszkoreit, J.; Jones, L.; Gomez, A.~N.; Kaiser, {\L}.; and Polosukhin, I. 2017.
\newblock Attention is all you need.
\newblock \emph{Advances in neural information processing systems}, 30.

\bibitem[{Wang et~al.(2024)Wang, Liu, Fan, Zhao, Kini, Yadav, Wang, Wen, Tang, and Liu}]{59}
Wang, H.; Liu, X.; Fan, W.; Zhao, X.; Kini, V.; Yadav, D.; Wang, F.; Wen, Z.; Tang, J.; and Liu, H. 2024.
\newblock Rethinking large language model architectures for sequential recommendations.
\newblock \emph{arXiv preprint arXiv:2402.09543}.

\bibitem[{Wang et~al.(2020)Wang, Louca, Hu, Cellier, Caverlee, and Hong}]{37}
Wang, J.; Louca, R.; Hu, D.; Cellier, C.; Caverlee, J.; and Hong, L. 2020.
\newblock Time to Shop for Valentine's Day: Shopping Occasions and Sequential Recommendation in E-commerce.
\newblock In \emph{Proceedings of the 13th International Conference on Web Search and Data Mining}, 645--653.

\bibitem[{Wang et~al.(2019)Wang, Hu, Wang, Cao, Sheng, and Orgun}]{35}
Wang, S.; Hu, L.; Wang, Y.; Cao, L.; Sheng, Q.~Z.; and Orgun, M. 2019.
\newblock Sequential recommender systems: challenges, progress and prospects.
\newblock \emph{arXiv preprint arXiv:2001.04830}.

\bibitem[{Wang, He, and Zhu(2024)}]{3}
Wang, Y.; He, X.; and Zhu, S. 2024.
\newblock EchoMamba4Rec: Harmonizing Bidirectional State Space Models with Spectral Filtering for Advanced Sequential Recommendation.
\newblock \emph{arXiv preprint arXiv:2406.02638}.

\bibitem[{Wang et~al.(2023)Wang, Lam, Wong, Liu, Zhao, Wang, Chen, Guo, and Tang}]{50}
Wang, Y.; Lam, H.~T.; Wong, Y.; Liu, Z.; Zhao, X.; Wang, Y.; Chen, B.; Guo, H.; and Tang, R. 2023.
\newblock Multi-task deep recommender systems: A survey.
\newblock \emph{arXiv preprint arXiv:2302.03525}.

\bibitem[{Yang et~al.(2024)Yang, Li, Zhao, Wang, Ma, Ma, Ren, Zhang, Xin, Chen et~al.}]{4}
Yang, J.; Li, Y.; Zhao, J.; Wang, H.; Ma, M.; Ma, J.; Ren, Z.; Zhang, M.; Xin, X.; Chen, Z.; et~al. 2024.
\newblock Uncovering Selective State Space Model's Capabilities in Lifelong Sequential Recommendation.
\newblock \emph{arXiv preprint arXiv:2403.16371}.

\bibitem[{Yoon and Jang(2023)}]{44}
Yoon, J.~H.; and Jang, B. 2023.
\newblock Evolution of deep learning-based sequential recommender systems: from current trends to new perspectives.
\newblock \emph{IEEE Access}, 11: 54265--54279.

\bibitem[{Yu, Mahoney, and Erichson(2024)}]{21}
Yu, A.; Mahoney, M.~W.; and Erichson, N.~B. 2024.
\newblock There is HOPE to Avoid HiPPOs for Long-memory State Space Models.
\newblock \emph{arXiv preprint arXiv:2405.13975}.

\bibitem[{Yuan et~al.(2019)Yuan, Karatzoglou, Arapakis, Jose, and He}]{42}
Yuan, F.; Karatzoglou, A.; Arapakis, I.; Jose, J.~M.; and He, X. 2019.
\newblock A simple convolutional generative network for next item recommendation.
\newblock In \emph{Proceedings of the twelfth ACM international conference on web search and data mining}, 582--590.

\bibitem[{Zhang et~al.(2023)Zhang, Chen, Zhao, Han, and Li}]{53}
Zhang, C.; Chen, R.; Zhao, X.; Han, Q.; and Li, L. 2023.
\newblock Denoising and prompt-tuning for multi-behavior recommendation.
\newblock In \emph{Proceedings of the ACM Web Conference 2023}, 1355--1363.

\bibitem[{Zhang et~al.(2022)Zhang, Du, Zhao, Han, Chen, and Li}]{54}
Zhang, C.; Du, Y.; Zhao, X.; Han, Q.; Chen, R.; and Li, L. 2022.
\newblock Hierarchical item inconsistency signal learning for sequence denoising in sequential recommendation.
\newblock In \emph{Proceedings of the 31st ACM International Conference on Information \& Knowledge Management}, 2508--2518.

\bibitem[{Zhang, Wang, and Zhao(2024)}]{25}
Zhang, S.; Wang, M.; and Zhao, X. 2024.
\newblock GLINT-RU: Gated Lightweight Intelligent Recurrent Units for Sequential Recommender Systems.
\newblock \emph{arXiv preprint arXiv:2406.10244}.

\bibitem[{Zhang and Sabuncu(2018)}]{32}
Zhang, Z.; and Sabuncu, M. 2018.
\newblock Generalized cross entropy loss for training deep neural networks with noisy labels.
\newblock \emph{Advances in neural information processing systems}, 31.

\bibitem[{Zhao et~al.(2023{\natexlab{a}})Zhao, Liu, Cai, Zhao, Liu, Zheng, Jiang, and Gai}]{39}
Zhao, K.; Liu, S.; Cai, Q.; Zhao, X.; Liu, Z.; Zheng, D.; Jiang, P.; and Gai, K. 2023{\natexlab{a}}.
\newblock KuaiSim: A comprehensive simulator for recommender systems.
\newblock \emph{Advances in Neural Information Processing Systems}, 36: 44880--44897.

\bibitem[{Zhao et~al.(2021)Zhao, Mu, Hou, Lin, Chen, Pan, Li, Lu, Wang, Tian et~al.}]{30}
Zhao, W.~X.; Mu, S.; Hou, Y.; Lin, Z.; Chen, Y.; Pan, X.; Li, K.; Lu, Y.; Wang, H.; Tian, C.; et~al. 2021.
\newblock Recbole: Towards a unified, comprehensive and efficient framework for recommendation algorithms.
\newblock In \emph{proceedings of the 30th acm international conference on information \& knowledge management}, 4653--4664.

\bibitem[{Zhao et~al.(2023{\natexlab{b}})Zhao, Wang, Zhao, Li, Zhou, Yin, Li, Tang, and Guo}]{49}
Zhao, X.; Wang, M.; Zhao, X.; Li, J.; Zhou, S.; Yin, D.; Li, Q.; Tang, J.; and Guo, R. 2023{\natexlab{b}}.
\newblock Embedding in recommender systems: A survey.
\newblock \emph{arXiv preprint arXiv:2310.18608}.

\bibitem[{Zhu et~al.(2024)Zhu, Liao, Zhang, Wang, Liu, and Wang}]{17}
Zhu, L.; Liao, B.; Zhang, Q.; Wang, X.; Liu, W.; and Wang, X. 2024.
\newblock Vision mamba: Efficient visual representation learning with bidirectional state space model.
\newblock \emph{arXiv preprint arXiv:2401.09417}.

\end{thebibliography}

\section{Appendix} \label{sec:ana}
\subsection{A. Complexity Analysis} In this section, we will analyze the time complexity of our proposed SIGMA framework. 
We denote training iteration as ${I}$, the sequence length as ${N}$, the embedding size as ${D}$ and the kernel size of Conv1d as $k$. Following the previous formulas, we analyze the time complexity of the key components: (i) Firstly, since our DS Gate consists of several linear layers and one Conv1d Layer, the computation cost of it can be calculated as $\mathcal{O}\left(3 \times N \times D^2 + N\times D \times k\right)$. (ii) Secondly, the time complexity of Mamba is known as $\mathcal{O}\left(N \times D\right)$~\cite{1}. In the PF-Mamba, we add one more reverse direction and several linear layers, therefore, the total complexity can be calculated as $\mathcal{O}\left(3 \times N \times D^2 + N\times D \times (k+2)\right)$. (iii) Thirdly, for the FE-GRU, which consists of Conv1d and GRU cell, the computational complexity can be calculated by simply adding GRU and Conv1d together: $\mathcal{O}\left(N \times D^2 + N \times D \times k\right)$. In conclusion, the whole time complexity of SIGMA is $\mathcal{O}\left(13 \times N \times D^2 + N\times D \times (2k+2)\right)$.  Considering the extreme situation when dealing with quite long sequences ($\left(N \gg D\right)$), it can be further simplified to $\mathcal{O}\left(N\right)$, compared to the $\mathcal{O}\left({N}^{2}\right)$ complexity of transformer-based models~\cite{7}, showing its superiority in efficiency, which also support by the experimental results on ML-1M dataset.
\subsection{B. Dataset Information}
We mainly evaluate our framework on five real-world datasets, \textit{i.e.,} Beauty, Sports, Games, Yelp and ML-1M, which are all large-scale public datasets that have been widely used as benchmarks in the next-item prediction task~\cite{35}.
\begin{itemize}
    \item \textbf{Yelp}\footnote{\url{https://www.yelp.com/dataset}}: This dataset is released by Yelp as part of their Dataset Challenge. The data source includes user reviews, business information, and user interactions on Yelp. It contains over $6.9$ million reviews, details on more than $150,000$ businesses, and user interaction data like check-ins and tips.
    \item \textbf{Amazon based}\footnote{\url{https://cseweb.ucsd.edu/jmcauley/datasets.html\#amazon_reviews}}: These three datasets, provided by Amazon, include customer reviews and metadata from the Beauty, Sports, and Games categories which feature millions of reviews, with attributes such as review text, ratings, product IDs, and reviewer information.
    \item \textbf{MovieLens-1M}\footnote{\url{https://grouplens.org/datasets/movielens/}}: The MovieLens 1M dataset is released by GroupLens Research. It comprises 1 million movie ratings from $6,041$ users on $3417$ movies with attributes like user demographics, movie titles, genres, and timestamps.
\end{itemize}

In the experiments, we employ a leave-one-out method for splitting the datasets and arrange all the user interactions sequentially by time. Moreover, for each user and item, we construct an interaction sequence by simply sorting their interaction records based on timestamps and ratings. Considering the average length and total samples of each dataset vary, we filter users and items with less than five recorded interactions for ML-1M, Beauty, and Games, following the setting in original papers~\cite{3}. For Yelp and Sports, we also added upper bounds ($100$ for Yelp and $200$ for Sports). 
\subsection{C. Evaluation Metrics}
In this section, we will detail the information and calculations of our selected evaluation metrics.
\begin{itemize}
    \item \textbf{HR@10}: HR@10 represents Hit Rate truncated at 10, which measures the fraction of users for whom the correct item is ranked within the top 10 predictions. Specifically, for a user $u$, it is calculated as:
    \begin{equation}
    \text{HR@10} = \frac{1}{|\mathcal{U}|} \sum_{u \in \mathcal{U}} \mathbb{1}\left( \text{rank}_{\text{correct}}(v_u) \leq 10 \right)
    \end{equation}
    where \( v_u \) is the correct item for user \( u \) and \( \mathbb{1}(\cdot) \) is an indicator function that equals 1 if the condition is true and 0 otherwise.

    \item \textbf{NDCG@10}: NDCG@10 represents Normalized Discounted Cumulative Gain truncated at 10 which evaluates the ranking quality by giving higher importance to items ranked at the top. It is defined as:
    \begin{equation}
    \begin{aligned}
    \text{IDCG@10} &= \sum_{k=1}^{10} \frac{2^{\text{rel}^\ast(v_{uk})} - 1}{\log_2(k+1)}\\
    \text{NDCG@10} &= \frac{1}{|\mathcal{U}|} \sum_{u \in \mathcal{U}} \frac{1}{\text{IDCG@10}(v_u)} \sum_{k=1}^{10} \frac{2^{\text{rel}(v_{uk})} - 1}{\log_2(k+1)} 
    \end{aligned}
    \end{equation}
    where  $\text{rel}(v_{uk})$ represents the relevance score of item $v_{uk}$ for user $u$, $\text{rel}^{*}(v_{uk})$ represents the ideal relevance score of the k-th item in the list, assuming that the most relevant items are ranked highest. And $\text{IDCG@10}(v_u)$ is the ideal DCG, which is the maximum possible DCG@10 for the correct item $v_u$.

    \item \textbf{MRR@10}: MRR@10 represents Mean Reciprocal Rank truncated at 10, which measures the average rank position of the first relevant item. For user $u$, it is defined as:
    \begin{equation}
    \text{MRR@10} = \frac{1}{|\mathcal{U}|} \sum_{u \in \mathcal{U}} \frac{1}{\text{rank}_{\text{correct}}(v_u)}
    \end{equation}
    where $\text{rank}_{\text{correct}}(v_u)$ is the rank position of the correct item $v_u$ in the top 10 predictions for user $u$.
\end{itemize}

\subsection{D. Implementation Details }
In this section, we present the details about the implementation of our SIGMA framework. The corresponding code can be found in the \textbf{Supplementary Materials}.
The Mamba block is one of the most important components of the SSM-based model (including SIGMA). So for fair comparison, we set the SSM state expansion factor to 32, the local convolution width to 4, and the block expansion factor to 2 for all models that include Mamba blocks. For the number of stacked layers, we set it to the defaulted 2 for all the selected RNN-based, transformer-based models and Mamba4Rec, comparing them with ECHO and our SIGMA with 1 layer.
Moreover, to address the sparsity of Amazon datasets and Yelp dataset, a dropout rate of 0.3 is used, compared to 0.2 for MovieLens-1M. 
\subsection{E. Baselines}
(a) \textbf{GRU4Rec}~\cite{11}: GRU4Rec utilizes Gated Recurrent Units (GRUs) to capture sequential dependencies within user interaction data. It is particularly effective for session-based recommendations, allowing the model to focus on the most recent and relevant user interactions to make accurate predictions. The model is known for its efficiency in handling session-based data, particularly in capturing user intent during a session~\cite{12}.
(b) \textbf{BERT4Rec}~\cite{15}: BERT4Rec adapts the BERT (Bidirectional Encoder Representations from Transformers) architecture for personalized recommendations. Unlike traditional sequential models, BERT4Rec considers both past and future contexts of user behavior by employing a bidirectional self-attention mechanism, which enhances the representation of user interaction sequences. This allows the model to predict items in a more context-aware manner.
(c) \textbf{SASRec}~\cite{8}: SASRec applies a multi-head self-attention mechanism to capture both long-term and short-term user preferences from interaction sequences. It constructs user representations by focusing on relevant parts of the interaction history, thereby improving the quality of recommendations, especially in scenarios where user preferences vary over time.
(d) \textbf{LinRec}~\cite{24}: LinRec simplifies the computational complexity of the traditional transformer models by modifying the dot product in the attention mechanism. This reduction in complexity makes LinRec particularly suitable for large-scale recommendation tasks where efficiency is crucial, without significantly sacrificing performance.
(e) \textbf{FEARec}~\cite{23}: FEARec enhances traditional attention mechanisms by incorporating information from the frequency domain. This hybrid approach allows the model to better capture periodic patterns and long-range dependencies in user interaction sequences, leading to more powerful and accurate recommendations.
(f) \textbf{Mamba4Rec}~\cite{2}: Mamba4Rec leverages Selective State Space Models (SSMs) to address the effectiveness-efficiency trade-off in sequential recommendation. By using SSMs, Mamba4Rec efficiently handles long behavior sequences, capturing complex dependencies while maintaining low computational costs. It outperforms traditional self-attention mechanisms, especially in scenarios involving long user interaction histories.
(g) \textbf{EchoMamba4Rec}~\cite{3}: Building on Mamba4Rec, EchoMamba4Rec introduces a frequency domain filter to remove noise and enhance the signal in sequential data. This bi-directional model processes sequences both forward and backward, providing a more comprehensive understanding of user behavior. The Fourier Transform-based filtering further improves the model's robustness and accuracy in predicting user preferences.

\subsection{F. Efficiency Comparison}
In this section, We will present and analyze the efficiency comparison on other datasets with our proposed SIGMA.
From Table~\ref{table: effi}, we can see that the SSM-based models (including our SIGMA) show consistent efficiency in the other three datasets (Yelp, Sports, and ML-1M). Although for ML-1M, our SIMGA shows an increase in GPU Memory due to our partial flipping method, which means we need to store and compute a new sequence for the reverse direction, our method still achieves higher efficiency remarkably compared with the transformer-based methods, except for LinRec. The experimental results present the fact that our SIGMA can achieve a better efficiency-effectiveness trade-off.
\begin{table}[h]
	\centering
	\caption{Efficiency Comparison: Inference time (ms) per batch and GPU memory (GB).}
 \label{table: effi}
	\begin{tabular}{cccc}
		\toprule[1pt]
		Dataset &	Model	&Infer.	&GPU Memory \\
		\midrule
\multirow{6}{*}{Yelp}
            &SASRec	&443ms	&9.28G\\
	    &FEARec	&483ms	&10.01G\\
		&LinRec	&\underline{353ms}	&\underline{}{7.46G}\\
            &Mamba & 361ms & \textbf{7.32G}\\
		&ECHO	&368ms	&\underline{8.46G}\\
		&SIGMA	&\textbf{352ms}	&8.27G\\\midrule
		\multirow{6}{*}{Sports}	
            &SASRec	&398ms	&7.27G\\
		&FEARec	&416ms	&7.84G\\
		&LinRec	&{294ms}	&{6.07G}\\
            &Mamba &\underline{293ms}&\textbf{5.89G}\\
		&ECHO	&291ms	&{5.99G}\\
		&SIGMA	&\textbf{283ms}	&6.03G\\\midrule
            \multirow{6}{*}{ML-1M}    
            & SASRec & 87ms & 21.51G \\
            & FEARec & 109ms & 21.28G \\
            & LinRec & \underline{59ms} & 11.79G \\
            & Mamba & \textbf{55ms} & \textbf{6.89G} \\
            & ECHO & 63ms & {9.92G} \\
            & SIGMA & \underline{59ms} & \underline{9.89G} \\
  \bottomrule[1pt]
	\end{tabular}
	\begin{flushleft}
	\end{flushleft}
\end{table}

\begin{figure*}[!t]
	\centering
	\includegraphics[width = \linewidth]{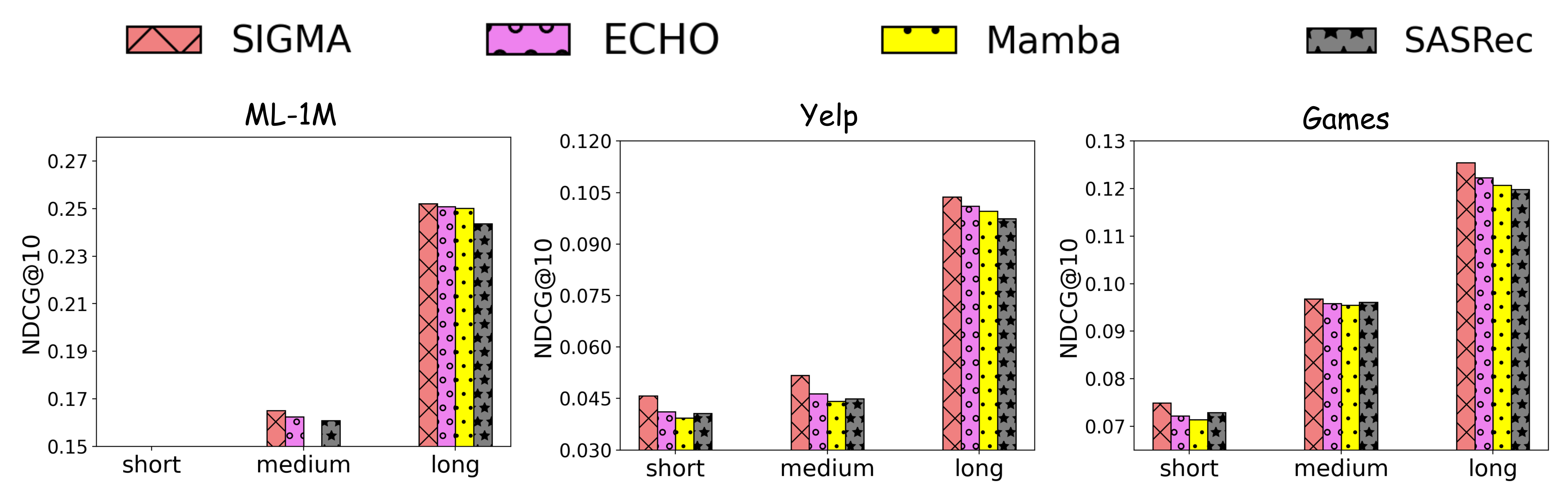}
	\caption{Grouped Users Analysis on ML-1M, Yelp and Games.}
        \label{fig:ap_lt}
\end{figure*}
\subsection{G. Grouped Users Analysis}
In this section, we will present the grouped user analysis on the other three datasets \textit{i.e.,} ML-1M, Yelp and Games.
We also group the user by their interaction length, followed by our experiment setting. The detailed distribution is listed in Table~\ref{table: distribution}. The performance of the selected models is presented in Figure~\ref{fig:ap_lt}. Noted that, as illustrated in Table~\ref{table: distribution}, the interaction length of users in ML-1M are all longer than 5, so the performance for the ``short" group in ML-1M is reasonably empty. All the experiments above further prove the superiority of our proposed SIGMA framework on all groups, showing the effectiveness of our FE-GRU and PF-Mamba in dealing with context modeling and short sequence modeling, respectively.

\begin{table}[h]
	\centering
	\caption{User sample distribution}
        \label{table: distribution}
	\begin{tabular}{cccc}
		\toprule[1pt]
		Dataset	&	Short(0-5) &	Medium(5-20)	&Long(20-inf)\\
		\midrule
		Games&		27631& 25084& 2426	\\
            Yelp&		34659&	42934&	5306\\
            ML-1M&		0&	177&	5863\\
		\bottomrule[1pt]
	\end{tabular}
\end{table}

\subsection{H. Ablation Study}
In this section, we will analyze the efficacy of three key components with SIGMA on other datasets \textit{i.e.,} Sports, Games, Yelp and ML-1M. According to the data statistics in the experiment setting, we can clearly see that Yelp, Sports, Beauty, and Games have similar average interaction lengths. Correspondingly, the ablation study on Sports, Games, and Yelp datasets shows a similar tendency to the one on Beauty. But for ML-1M, due to the relatively long sequence, we can find that FE-GRU contributes the least since it mainly focuses on enhancing the hidden representation for short sequences, while the other two components show a remarkable drop when removing them, proving the effectiveness of our designed PF-Mamba in contextual information modeling.
\begin{table}[h]
	\centering
	\caption{Ablation study on other datasets.}
 \label{table: effi}
	\resizebox{.5\textwidth}{!}{%
	\begin{tabular}{ccccc}
		\toprule[1pt]
		Dataset &	Methods	&HR@10	&NDCG@10& MRR@10 \\
		\midrule
\multirow{4}{*}{Sports}
            &Default	&\textbf{0.0735}	&\textbf{0.0590} &\textbf{0.0556}\\
            &w/o partial flipping &0.0711	&0.0582 &0.0544\\
		&w/o DS gate	&0.0713	&0.0577 &0.0529\\
		&w/o FE-GRU	&0.0622	&0.0481 &0.0473\\\midrule
		\multirow{4}{*}{Games}	
            &Default	&\textbf{0.1627}	&\textbf{0.1088} &\textbf{0.0924}\\
            &w/o partial flipping &0.1601	&0.1067 &0.0911\\
		&w/o DS gate	&0.1597	&0.1054 & 0.0889\\
		&w/o FE-GRU	&0.1562	&0.0978 &0.0861\\\midrule
            \multirow{4}{*}{Yelp}    
            &Default	&\textbf{0.0629}	&\textbf{0.0412} &\textbf{0.0346}\\
            &w/o partial flipping &0.0611	&0.0402 &0.0327\\
		&w/o DS gate	&0.0613	&0.0389 &0.0314\\
		&w/o FE-GRU	&0.0595	&0.0377 &0.0301\\\midrule
            \multirow{4}{*}{ML-1M}    
            &Default	&\textbf{0.3308}	&\textbf{0.1906} &\textbf{0.1479}\\
            &w/o partial flipping &0.3289	&0.1887 &0.1459\\
		&w/o DS gate	&0.3288	&0.1866 &0.1443\\
		&w/o FE-GRU	&0.3304	&0.1901 &0.1471\\
  \bottomrule[1pt]
	\end{tabular}}
	\begin{flushleft}
	\end{flushleft}
\end{table}
\subsection{I. Guideline for reproducibility}
In this section, we will provide detailed guidelines for reproducing our results.
In the model file we released, we have \textbf{baseline} and \textbf{baseline\_config} folders, which respectively contain almost all the baseline methods (others can be easily found in recbole's original models) and corresponding configurations mentioned in the overall experiment; \textbf{datasets} folder storing the chosen datasets (Beauty, Sports, Games, Yelp and ML-1M) in atomic files to fit the recbole framework~\cite{30}; \textbf{model} file containing the \textbf{gated\_mamba.py}, which is the main structure of our SIGMA. To reproduce our proposed SIGMA framework, you can directly run the \textbf{run.py} file with the proper command. It is noteworthy that the environment and version of \textbf{mamba\_ssm} are quite important in reproducing the experimental results, so please check your environment referencing the \textbf{requirements.yaml} file.
Specifically, you are recommended to run \textbf{run.py} for successful reproduction, which contains the calls to \textbf{gated\_mamba.py}. For the relevant parameters to perform the experiments in hyperparameter analysis, you can directly set it in \textbf{config.yaml} with other dataset settings and model settings. We also attach the \textbf{RecMamba.ipynb} to show the raw training procedure in Colab.
\clearpage
\end{document}